\documentclass[lettersize,journal]{IEEEtran}
\usepackage{amsmath,amsfonts}
\usepackage{algorithmic}
\usepackage{algorithm}
\usepackage{array}
\usepackage[caption=false,font=normalsize,labelfont=sf,textfont=sf]{subfig}
\usepackage{textcomp}
\usepackage{stfloats}
\usepackage{url}
\usepackage{verbatim}
\usepackage{graphicx}
\usepackage{cite}
\usepackage{booktabs}
\usepackage{multirow}
\usepackage{threeparttable}

\begin{document}

\title{Two-person Graph Convolutional Network for Skeleton-based Human Interaction Recognition}

\author{Zhengcen~Li, Yueran~Li, Linlin~Tang, Tong~Zhang, and Jingyong~Su

        \thanks{Corresponding author: Jingyong Su (e-mail: sujingyong@hit.edu.cn)}
        \thanks{Zhengcen Li, Yueran Li, Linlin Tang and Jingyong Su are with the Harbin Institute of Technology (Shenzhen), Shenzhen 518055, China}
        \thanks{Tong Zhang and Jingyong Su are with the Peng Cheng Laboratory, Shenzhen 518055, China}
        \thanks{Code is available on-line at https://github.com/mgiant/2P-GCN}
}

\markboth{Journal of IEEE Transactions on Circuits and Systems for Video Technology}%
{Li \MakeLowercase{\textit{et al.}}: Two-person Graph Convolutional Network for Skeleton-based Human Interaction Recognition}


\maketitle

\begin{abstract}
    Graph convolutional networks (GCNs) have been the predominant methods in skeleton-based human action recognition, including human-human interaction recognition. However, when dealing with interaction sequences, current GCN-based methods simply split the two-person skeleton into two discrete graphs and perform graph convolution separately as done for single-person action classification. Such operations ignore rich interactive information and hinder effective spatial inter-body relationship modeling. To overcome the above shortcoming, we introduce a novel unified two-person graph to represent inter-body and intra-body correlations between joints. Experiments show accuracy improvements in recognizing both interactions and individual actions when utilizing the proposed two-person graph topology. In addition, We design several graph labeling strategies to supervise the model to learn discriminant spatial-temporal interactive features. Finally, we propose a two-person graph convolutional network (2P-GCN). Our model achieves state-of-the-art results on four benchmarks of three interaction datasets: SBU, interaction subsets of NTU-RGB+D and NTU-RGB+D 120.
\end{abstract}


\section{Introduction}
\label{sec:intro}
    Human-human interaction recognition is an increasingly important task in computer vision, which aims at identifying two-person interactive activities from videos and mainly benefits content-based retrieval, video surveillance, and human-robot interaction~\cite{Ji2020survey}. Conventional solutions to activity detection and classification often consist of video processing techniques and machine learning approaches~\cite{khaire2022survey}. Compared to regular RGB videos, the skeleton representation, which depicts the human body as 3D key-point coordinates, is computationally less expensive and is more robust against occlusions and background changes. Moreover, skeleton capturing devices~\cite{zhang2012kinect} and pose estimation algorithms~\cite{cao2019openpose,sun2019hrnet} provide economical ways to obtain accurate joint coordinates. Therefore, skeleton-based action recognition has been extensively studied in recent years~\cite{yun2012SBU,vemulapalli2014liegroup,shahroudy2016ntu,Hou2018SOS,yan2018ST-GCN,perez2021LSTM-IRN}.
    
    
    Despite significant progress has been made in general skeleton-based action recognition, it remains difficult to extract skeleton data of multiple people from videos and recognize their interactions~\cite{stergiou2019survey,khaire2022survey}. This paper focuses on a simple yet essential landmark, skeleton-based human-human interaction recognition.
    
    Existing interaction recognition methods can be categorized into two groups: hand-crafted approaches and deep learning models. Traditional methods~\cite{zhang2012spatio,kong2012learning,yun2012SBU,ji2014interactive} mostly start with specified prior knowledge and utilize typical feature representations; they often use the support vector machine (SVM) to group transformed vectors. More recent solutions~\cite{donahue2015long,sadegh2017encouraging,wang2016hierarchical,perez2021LSTM-IRN,men2021two} use a recurrent neural network (RNN) or convolution neural network (CNN) to extract information from joint coordinates, motion patterns, or features processed using hand-crafted techniques. These models cannot directly process non-Euclidean skeleton data. Therefore, they often need specific preprocessing for target datasets to achieve competitive performance.
    \begin{figure}[tp]
        \centering
        \includegraphics[width=0.48\textwidth]{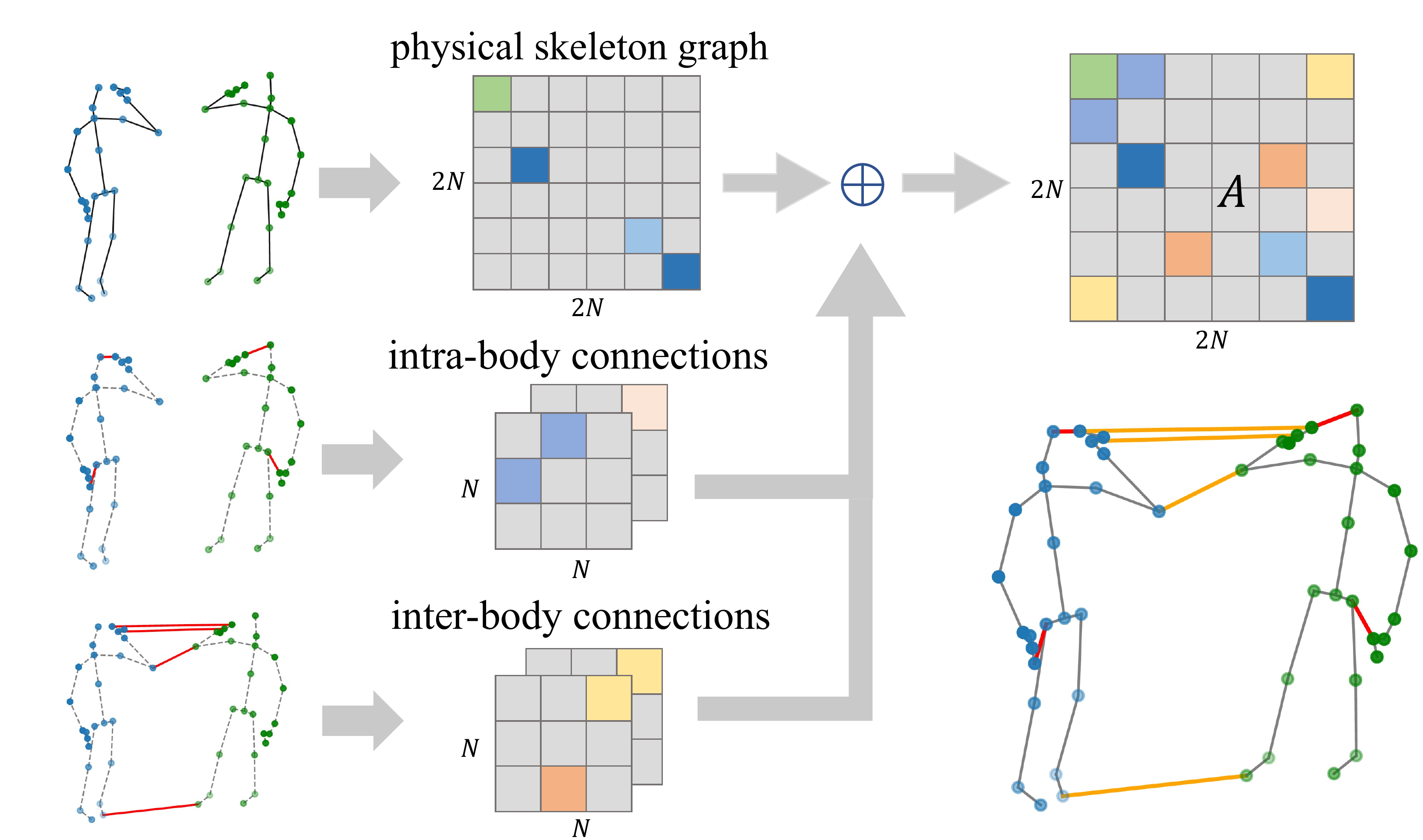}
        \caption{Illustration of the proposed two-person graph and its corresponding adjacency matrix, where $N$ is the number of joints for each person.}
        \label{Fig:relation_matrix}
    \end{figure}
    
    Recently, graph convolutional networks have been introduced in skeleton-based action recognition. Numerous studies~\cite{yan2018ST-GCN,shi20192sAGCN,song2021RAGCN,song2020ResGCN,chen2021CTR-GCN,zhu2021DRGCN,liu2020MS-G3D} have shown the advantage of GCN in extracting spatial-temporal features from graph-structured skeleton data. However, when handling interaction sequences, most GCN-based methods simply split two-person skeletons apart in each frame and consider them as two isolated individuals~\cite{yan2018ST-GCN,shi20192sAGCN,liu2020MS-G3D,song2020ResGCN,chen2021CTR-GCN}. Crucial interactive information might be lost under this processing. 
    
    In this work, we introduce a novel two-person graph to represent human-human interactions. As shown in Fig.~\ref{Fig:relation_matrix}, in each frame, an interaction sequence is represented by a two-person graph as a whole. The relational adjacency matrix of the two-person graph can simultaneously depict the intra-body correlations which are far away in the physical skeleton graph and inter-body relations between two people's joints. To effectively model these intra-body and inter-body dependencies, a series of graph edge labeling strategies are introduced to denote how these edges are connected and their correlation strength. With the proposed two-person graph structure, GCN can extract spatial inter-body and intra-body features within a single graph convolution operation.
    

    The proposed two-person graph does not increase computation costs and model parameters compared to two stacked single-person graphs. Furthermore, the two-person graph can convey more information and has good scalability. If there is only one person in the scene, a two-person graph could be more informative. For instance, to reflect individual actions that are part of interactions (e.g., making phone calls), we can create a mirror skeleton as an imagined second person. Experimental results show that the two-person graph representation also improves the classification accuracy of individual actions. In addition, a two-person graph can be easily generalized to multi-person scenarios. We can further increase the graph scale and build a multi-person graph that captures all the people we are interested in for better describing group activities like football games. In experiments, we apply the proposed two-person graph in several GCNs and notice that their performances have been significantly improved. Thus, the two-person graph representation is an ideal substitution for the commonly used single-person graph.
    
    Based on the two-person graph, we design a two-person graph convolution block (2P-GCB) and propose 2P-GCN for skeleton-based interaction recognition.
    Experiments conducted on three public human-human interaction datasets~\cite{yun2012SBU,shahroudy2016ntu,liu2019ntu120} show that the proposed 2P-GCN outperforms existing skeleton-based interaction recognition approaches. 
    Our contributions can be summarized as follows.
    \begin{itemize}
   	    \item We introduce the two-person graph representation as a substitution for the commonly-used single-person graph and design a series of edge labeling strategies for the two-person graph. Our two-person graph unifies the notation of inter-body and intra-body correlations. We conduct experiments under several GCNs. For all evaluated models, two-person graph obtains higher accuracy than their original graphs on all benchmark datasets. 
    	
    	\item We develop the 2P-GCB, comprising of a two-person spatial graph convolutional layer, a multi-scale temporal convolutional layer, and a spatial-temporal part attention module. Our 2P-GCB enables GCN to learn discriminant spatial-temporal interactive information from two-person interactions. 
 
    	\item We propose 2P-GCN for skeleton-based interaction recognition. We evaluate our model on the interaction subset of NTU-RGB+D~\cite{shahroudy2016ntu} and NTU-RGB+D 120~\cite{liu2019ntu120}. Our model achieves state-of-the-art performances on all four benchmarks.
    \end{itemize}

\section{Related work}
    \subsection{Traditional Human Interaction Recognition}
    Human interaction recognition is a subtopic of action recognition. Many action recognition datasets ~\cite{kuehne2011hmdb51,shahroudy2016ntu,kay2017kinetics400,liu2019ntu120} contain both individual actions and multi-person interactions. 
    
    Most previous studies~\cite{zhang2012spatio,kong2012learning,yun2012SBU,ji2014interactive} were based on handcrafted features from videos. They often chose a particular descriptor and then applied SVM to classify the extracted feature vectors. These methods can be divided into local feature approaches, which rely on detecting informative points in the video, and template-based approaches, which consider regions corresponding to a person's body parts~\cite{stergiou2019survey}. 
    More recent work~\cite{donahue2015long,sadegh2017encouraging,wang2016hierarchical} constructed a CNN- or RNN-based end-to-end network to extract features and perform classification, or combined deep learning models with handcrafted techniques. RGB videos are sensitive to irrelevant factors such as camera motion, background changes, lightness, and occlusions. Therefore, RGB-based methods often require additional preprocessing and higher computing resources.

    For higher accuracy and robustness, features should be invariant to camera viewpoints and background environments. Therefore, some researchers have focused on skeleton data. Yun et al.~\cite{yun2012SBU} derived interaction features from the geometric relations among different joints, including intra-body, inter-body, inter-frame, and intra-frame. Then, an SVM was adopted for the classification. Ji et al.~\cite{ji2014interactive} calculated the spatial-temporal joint features of eight interactive body parts named poselets. After removing the redundant information by employing contrast miming, a dictionary was generated based on these features. 
    Wu et al.~\cite{wu2017recognition} used sparse-group LASSO to select factors to deal with real-time interaction detection tasks automatically. Nguyen~\cite{nguyen2021geomnet} proposed a two-person interaction representation based on the Riemannian geometry of the underlying manifolds. They represented a 3D skeleton sequence by a set of symmetric positive definite matrices and introduced a neural network GeomNet for action classification problems.
    Perez et al.\cite{perez2021LSTM-IRN} proposed a relational reasoning network to learn from 3D coordinates of pair-wise joints and perform classification based on inferred relations. 
    
    Existing interaction recognition methods are often weak in effectively modeling spatial intra-body dependencies, motion patterns, and interactive features within a single network. Therefore, they may not obtain competitive results on large-scale datasets.
    
    
    \subsection{GCN-based Human Interaction Recognition}
    
    Spatial-based GCN models~\cite{song2020ResGCN,shi20192sAGCN,liu2020MS-G3D,chen2021CTR-GCN} have been the predominant approaches in skeleton-based action recognition. Most GCNs follow the feature aggregation rule of ~\cite{welling2016semi} and the graph convolution formulation of ~\cite{yan2018ST-GCN}. When dealing with interaction actions, conventional GCN-based models often consider only one key skeleton or split two-person skeletons into two separate ones.
    
    Several GCN-based methods have been used to model the interactive features~\cite{yang2020pairwise,zhu2021DRGCN,li2021K-GCN,gao2022AIGCN}. Yang et al.~\cite{yang2020pairwise} introduced a pairwise graph by manually adding inter-body links to the isolated two-person graph. Then, they extended ST-GCN to ST-GCN-PAM, which adopts pairwise graph convolution to learn the interactive relation between two bodies. Li et al.~\cite{li2021K-GCN}  introduced a shared knowledge-given graph and dynamically inferred knowledge-learned graphs to capture the relations between inter-body joints explicitly. These two graphs and the naturally connected graph are with three parallel branches in the spatial graph convolution block. Zhu et al.~\cite{zhu2021DRGCN} adopted a GCN with separate graphs but explicitly modeled the interactive features. They proposed inter-body graph convolution with a dynamic relational adjacency matrix to capture interaction and conducted it as a parallel operation of intra-body graph convolution. Gao et al.~\cite{gao2022AIGCN} designed attention-driven modules to capture two-person spatial interactive relationships with dynamic attention encoding. They considered the two-person skeletons as a whole and proposed the attention mechanism in both IAE-GCN and IAE-TCN for better spatial and temporal modeling.
    
    
    In~\cite{li2021K-GCN} and \cite{zhu2021DRGCN}, the inter-body and intra-body graph convolution are performed separately. These networks are designed to learn interactive information explicitly; however, they still follow the formulation of single-person graph convolution. Therefore, current GCN-based interaction recognition methods often lead to a complex network structure and weak scalability. Compared to the aforementioned GCNs, our method is based on two-person graph topology and attempts to explore a general and effective graph representation of interactive skeletons.

\section{Method}
\label{Sec:method}

\subsection{Two-person Graph}
\label{Sec:2p-graph}
    
    In skeleton representation, we use $N$ joints and several bones to denote one person's body. For one person in each frame, the skeleton can be represented as a spatial graph
    $\mathcal{G}=(\mathcal{V}, \mathcal{E}, \mathcal{X})$,
    where
    $\mathcal{V}=\{v_{i}|i=1,...,N\}$
    is the vertex set of $N$ joints, and $\mathcal{E}=\{e_{ij}|i,j=1,...N\}$ is the edge set. Undirected edge $e_{ij}\in\mathcal{E}$
    represents the bone connection between $v_i$ and $v_j$. The relational adjacency matrix $\mathbf{A}\in\mathbb{R}^{N\times N}$ denotes all edges and its element $a_{ij}$ denotes the correlation strength of $e_{ij}$. 
    $\mathcal{X}\in\mathbb{R}^{C\times{T}\times N}$
    is the $C$-channel feature map of a sequence in $T$ frames, typically the 3D joint coordinates. 
    
    For multi-person interactions, a skeleton sequence is represented as $\mathcal{X}\in\mathbb{R}^{C\times{T}\times M\times N}$, where $M$ is the number of people in the scene. In this work, $M=2$. Most previous studies have split the $M$ persons apart and generated $M$ single-person graphs for each sequence. In contrast, we consider the two people as a whole and organize the input skeleton data as $\mathcal{X}\in\mathbb{R}^{C\times{T}\times 2N}$ to model the inter- and intra-body connections with a single adjacency matrix $\mathbf{A}\in\mathbb{R}^{2N\times 2N}$.
    In physical word, a two-person skeleton is a fixed and disconnected graph with two components representing two person's bodies. To better capture the interactive information between faraway joints in the physical skeleton graph, we introduce two types of edges: 1) intra-body connections, including the virtual edges between joints of the same person. 2) inter-body connections, which link joints from two different bodies. These connections enable us to model the dependencies of interactive joints explicitly. In practice, these three types of edges are integrated into a two-person graph, as illustrated in Fig.~\ref{Fig:relation_matrix}. 

\subsection{Edge Labeling Strategies}
    We have discussed the three types of edges in the two-person graph. Note that the strategy for labeling the inter-body and intra-body edges can be defined arbitrarily and may dramatically influence the model performance. Therefore, in order to find out a suitable graph representation, we explore several labeling strategies based on the two-person graph.
    
    
    \begin{figure}[tbp]
        \centering
        \includegraphics[width=0.48\textwidth]{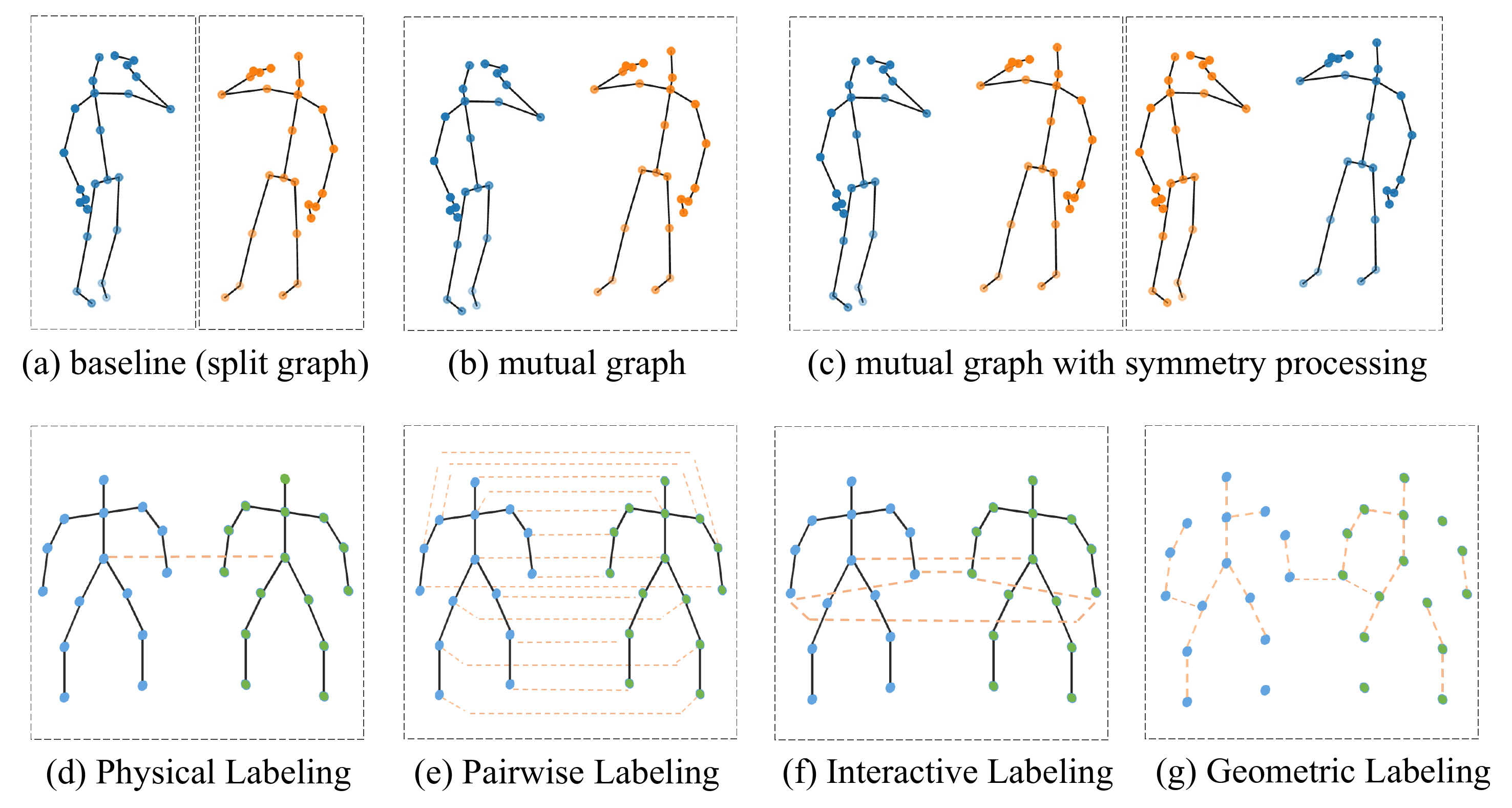}
        \caption{A brief visualization of different graph scales and edge labeling strategies. (a)-(c) are graph scales that decide the number of joint nodes and the number of graphs created for a sequence. The first person is in blue while the second person is in orange. (d)-(g) are labeling strategies, which depict the edges of the two-person graph.}
        \label{Fig:labeling}
    \end{figure}
    
\label{Sec:labeling}

    \paragraph{Physical Labeling}
    Physical labeling reflects the natural connection of the human body, which is the baseline strategy for our two-person graph. In addition, we manually add one edge between two body centers to keep the whole graph connected. 
    
    \paragraph{Pairwise Labeling}
    Yang et al.~\cite{yang2020pairwise} developed pairwise links in a two-person graph. This labeling strategy links every corresponding joint among the two-person skeletons (e.g, person A's left hand to person B's left hand). With pairwise edges, the GCN model can explicitly extract interactive correlations between inter-body joint pairs. 
    
    \paragraph{Interactive Labeling}
    In Interactive Labeling, we only connect joints that tend to have more interaction probability. In particular, we construct (1) two edges of the pairwise inter-body links between both hands and (2) two edges of the intra-body connections between one's left hand to right hand. Unlike pairwise labeling, interactive labeling would be more effective as it satisfies graph sparsity by filtering out less relevant correlations. In experiments, pairwise labeling and interactive labeling are integrated with the edges denoted by physical labeling.
    
    \paragraph{Geometric Labeling}
    The above labeling strategies construct a fixed graph based on our prior knowledge. If $v_i$ connects to $v_j$, then in adjacency matrix $\mathbf{A}$, $a_{ij}=1$. In contrast, current GCNs mostly adopt a dynamically inferred topology for a single-person graph and assign different weights for different edges. In multi-person scenarios, interaction occurs more frequently and could hide crucial information for recognition. Therefore, inspired by \cite{zhu2021DRGCN}, we utilize the spatial relationship as a metric of interaction. The edge strength is defined by the distance of joints in the 3D Euclidean space. The correlation between two joints is measured as
    \begin{equation}
    \label{Eq:geometric}
        a_{ij} = \delta(x_i,x_j)=
        \frac{1}{T}\sum_t \exp(-\frac{||(x^t_i-x^t_j)||^2}{C}) 
    \end{equation}
    
    where $a_{ij}$ is the element of adjacency matrix $\mathbf{A}$ and corresponds to the relation of joint $i$ and joint $j$. $x^t_i\in\mathbb{R}^C$ is a $C$-dimensional feature vector of the $i\in\{1,...2N\}$ th joint in frame $t\in\{1,...T\}$. We briefly visualize these labeling strategies in Fig.~\ref{Fig:labeling}.
    
    \paragraph{Adaptive Labeling}
    Recent GCN-based methods \cite{shi20192sAGCN,chen2021CTR-GCN} designed specialized modules to infer an adaptive graph topology. To examine whether the adaptive graph structure still works under a two-person graph, we introduce two adaptive labeling strategies which have been evaluated to be effective under a single-person graph topology. (1) AAGC~\cite{shi2020AAGCN}.
    In AAGC, the graph topology is dynamically learned and updated during training.
    (2) CTRGC~\cite{chen2021CTR-GCN} proposed a channel-wise topology refinement module to generate channel-specified adaptive graph topologies. It was proved to have stronger representation capability than the previous graph structures.

    Yan et al.~\cite{yan2018ST-GCN} proposed three graph partitioning strategies under a single-person physical graph. As for comparison, the partitioning strategy assigns neighbor nodes into different subsets based on a given graph; while the labeling strategy defines edges in the graph and their weights. 

    \begin{figure*}[t]
        \centering
        \includegraphics[width=\textwidth]{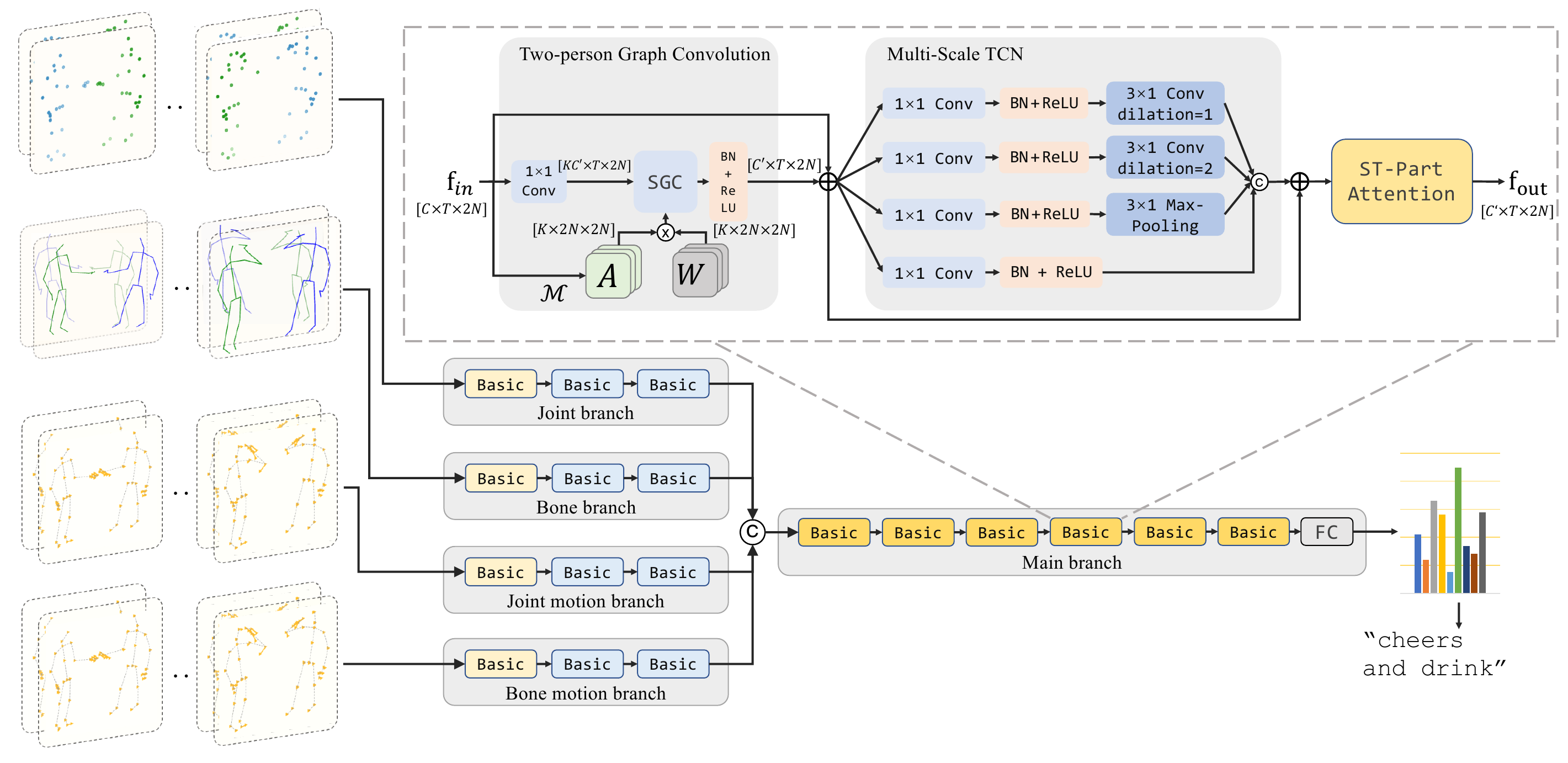}
        \caption{Illustration of the proposed 2P-GCN and detailed structure of the basic block, where $C,C',T,N$ and $K$ denote the numbers of input channels, output channels, frames, joints, and subsets in SGC, respectively. $\odot$, $\otimes$, and © represents the matrix production, element-wise production, and concatenation, respectively. $\mathcal{M}$ is the graph labeling strategy as discussed in Section \ref{Sec:labeling} and $\mathbf{A}$ denotes the corresponding adjacency matrix.}
        \label{Fig:2pgcn}
    \end{figure*}

\subsection{2P-GCN}
\label{Sec:2pgcn}

    We propose the two-person graph convolutional network for interaction recognition. First, the entire architecture is demonstrated. Then, the detailed implementations of the two-person spatial graph convolution (SGC) layer, multi-scale temporal convolution network (TCN), and spatial-temporal-part attention (ST-PartAtt) module are discussed in order.
    
    Current start-of-the-art methods usually apply multi-stream architecture, consisting of several identical GCN streams, to process different input data (e.g., joints, bones, motions) and fuse the prediction score at the last stage. Multi-stream fusion is an effective method to enhance the model performance\cite{shi20192sAGCN,chen2021CTR-GCN,zhu2021DRGCN}. However, the model parameters and computational costs grow linearly with the addition of inputs. Therefore, a multi-stream model often leads to higher complexity. Owing to this, following \cite{song2020ResGCN} and \cite{liu2020MS-G3D}, we construct four input branches--joint, bone, joint motion, and bone motion--each of which contains a few blocks. Detailed information on how we obtain each of these data types is in Sec.~\ref{Sec:preprocessing}. Then, we fuse them in the middle of the network by concatenating the results from all branches. With this architecture, our model can retain the rich information from different inputs but has significantly lower complexity than the aforementioned multi-stream methods.
    
    According to Yan et al.~\cite{yan2018ST-GCN}, the spatial graph convolution (SGC) is operated as 
    \begin{equation}
        \label{Eq:SGC1}
        f_{\text {out }}\left(v_{i}\right)=\sum_{v_{j} \in B\left(v_{ i}\right)} \frac{1}{Z_{i}\left(v_{j}\right)} f_{in}\left(v_{ j}\right) \cdot \mathbf{w}\left(l_{i}\left(v_{j}\right)\right),
    \end{equation}
    where $f_{in}()$ and $f_{out}()$ are input and output features of the corresponding joints, $v_{i}$ is the $i$-th joint node, $B(v_{i})$ is the neighbor set of $v_{i}$, the label function $l_{i}:B(v_{i})\to \{0,...,K-1\}$ maps a neighbor node to one of $K$ subsets, $\mathbf{w}$ is the weighting function, and $Z_{i}(v_{j})$ is introduced to balance the contributions of different subsets.
    
    In~\cite{yan2018ST-GCN}, the maximum graph sampling distance $D$ is set to $1$, which decides the number of nodes in $B(v_{i})$.
    In this paper, we increase $D$ to $2$ as the graph scale is doubled and maintain the number of subsets $K=3$ unchanged. Accordingly, we adopt the distance partitioning strategy in~\cite{yan2018ST-GCN}, where $l_{i}(v_{j})=d(v_{i},v_{j})$ and $d()$ is the graphic distance metric function. With the adjacency matrix $\mathbf{A}$, Eq.~\ref{Eq:SGC1} is implemented as
    \begin{equation}
        \label{Eq:SGC}
        \mathbf{f}_{out}=\sum_{d=0}^D \mathbf{\Lambda}_{d}^{-\frac{1}{2}} \mathbf{A}_{d} \mathbf{\Lambda}_{d}^{-\frac{1}{2}} \mathbf{f}_{in} \mathbf{W}_{d} ,
    \end{equation}
    where $\mathbf{f}_{in}$ and $\mathbf{f}_{out}$ are the input and output feature maps, $\mathbf{A}_d$ denotes the $d$-hop joint pairs in a two-person graph defined by one labeling strategy,
    $\mathbf{\Lambda}_d$ is introduced to normalize $\mathbf{A}_d$, and the learnable parameter $\mathbf{W}_d$ is introduced as edge importance weighting.
    
    In the temporal domain, we design a four-branch multi-scale temporal convolution network to extract temporal motion features in consecutive frames. As shown in the top-middle of Fig.~\ref{Fig:2pgcn}, each branch contains one $1\times1$ bottleneck block for channel reduction, one BatchNorm, and one ReLU layer, followed by a $3\times1$ convolutional layer with different dilations in the first two branches and a max-pooling layer in the third branch. With the two-person SGC layer and the multi-scale TCN, our network can extract and accumulate spatial and temporal interactive features simultaneously. 
    
\begin{figure}[tbp]
        \centering
        \includegraphics[width=0.48\textwidth]{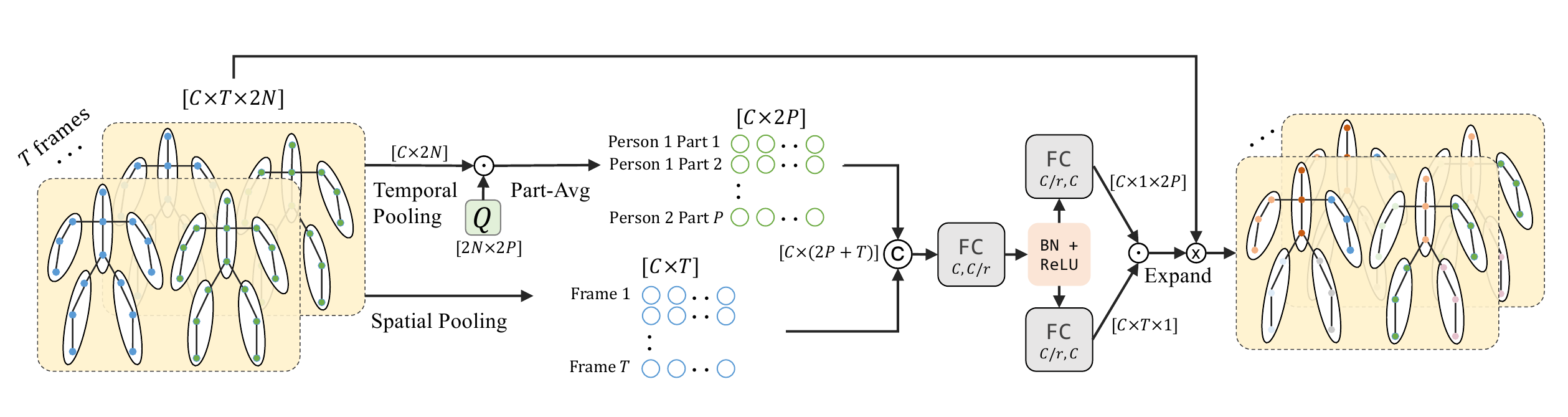}
        \caption{The structure of the proposed ST-Part Attention module, where $C,T$ and $N$ denote the numbers of input channels, frames, and joints, respectively. $r = 4$ is the channel reduction ratio, $Q$ is for part-level average pooling. $\odot$, $\otimes$, and © represent the matrix production, element-wise production, and concatenation, respectively.}
        \label{Fig:stpa}
    \end{figure}

    The attention mechanism is widely used in sequence modeling tasks. Inspired by \cite{song2020ResGCN,song2022Eff-GCN}, we design a spatio-temporal attention block, named ST-Part Attention. 
    The whole procedure is illustrated as Fig.~\ref{Fig:stpa}. Each human body has been divided into $P=5$ functional parts: both hands, both legs, and the torso. We first average the input feature map in the frame and part levels, respectively. Then, we concatenate the obtained feature vectors and feed them into a fully-connected layer, compressing them to a $C/r\times(2P+T)$ vector. Next, we adopt two fully-connected layers to obtain frame-wise and part-wise attention scores. Finally, we multiply these two attention scores and calculate the final spatial-temporal part attention map. In comparison, Part-Att\cite{song2020ResGCN} adopts global average pooling and produces a time-invariant attention map for each body part, while ST-Joint Att\cite{song2022Eff-GCN} generates an attention map for each joint in each frame. Since the computation is heavy for calculating the $T\times M\times N$ attention matrix for all joints in a multi-person graph, we propose ST-Part Attention as a lighter and more robust implementation.

\subsection{Data Pre-processing and Symmetry Processing}
\label{Sec:preprocessing}

    High-performance methods in general action recognition\cite{shi20192sAGCN,chen2021CTR-GCN,liu2020MS-G3D} usually propose a multi-stream network with various forms of input data. We follow these pre-processing ideas~\cite{song2020ResGCN,liu2020MS-G3D} and create four input features: joint data, bone data, joint motion, and bone motion.
    
    A raw sample in the dataset is a skeleton sequence represented as $\mathcal{X}=\mathbb{R}^{C\times T\times 2N}$, where $C,T,N$ denote the input coordinates, frames, and joints, respectively. In the preprocessing phase, for joint $i\in[1,..,2N]$ at frame $t\in[1,..,T]$, we expand the input data to $2C$ channels for each of the following branches. (1) \textbf{Joint}. The joint data are the concatenation of the raw skeleton $\mathcal{X}^{t}_i$ in first $C$ dimension and the relative coordinates of each person to the body center $\mathcal{X'}^{t}_i$ in the last $C$ dimension. (2) \textbf{Bone}. The bone vectors and angles between each bone to the axes form the bone features. The first $C$ dimension $\mathbf{l}^t_i\in \mathbb{R}^C$ is the bone vector pointing from itself toward its adjacent node that is closer to the body center. The bone angles of joint $i$ at time $t$ are calculated by
    \begin{equation}
        \mathbf{\Theta}_{i,c}^t = \arccos{\frac{\mathbf{l}_{i,c}^t}{\sqrt{||\mathbf{l}_{i}^t||^2}}},
    \end{equation}
    where $c\in\{x,y,z\}$ denotes the 3D coordinates for $C=3$ and $c\in\{x,y\}$ for $C=2$.
    (3) \textbf{Joint motion}. The joint motion data include the velocities $\mathbf{v_{joint}}$ and accelerations $\mathbf{a_{joint}}$. As in Eq.~\ref{Eq:joint-motion}, the velocity of joint $i$ at frame $t$ is the movement relative to itself in the previous frame, and the acceleration is the difference in the velocities in two adjacent frames. 
    \begin{equation}
        \label{Eq:joint-motion}
        \begin{split}
        \mathbf{v_{joint}}^t_i & = \mathcal{X}^{t}_i - \mathcal{X}^{t-1}_i, t = [2,...,T] \\
        \mathbf{a_{joint}}^t_i & = \mathbf{v_{joint}}^{t}_i - \mathbf{v_{joint}}^{t-1}_i, t = [2,...,T]
        \end{split}
    \end{equation}
    In particular, $\mathbf{v_{joint}}^1_i = \mathbf{a_{joint}}^1_i = \mathcal{X}^{1}_i$.
    (4) \textbf{Bone motion}. Bone motion is defined similarly as joint motion, with the only difference being that the input data are bone vectors instead of joint coordinates.
    
    To summarize, we augment the input feature of each type $\mathcal{X}=\mathbb{R}^{C\times T\times 2N}$ to a $2C\times T\times 2N$-dimensional tensor. Then, we design four identical input branches and feed the four input feature maps into them, respectively. Afterwards, we concatenate the output features of all four branches at a middle stage in our network and apply one main branch to process the fused features. 

    Besides, we notice that a number of interactions are not symmetric. For instance, shaking hands can be considered a symmetric action, where the roles of two individuals in the interaction activity are similar and interchangeable. However, interactions like kicking are not symmetric, as they always involve one subject and one object. During our experiments, we find that the order of placing each person in a multi-person graph considerably influences the model's accuracy. We propose two interpretations: the distribution imbalance caused by the train/test split of the dataset, or the inherent asymmetry of some interactions.
    
    To address this issue, we develop two solutions by reducing the high intra-class variation caused by the order of people. The first solution is based on the first hypothesis, in which we randomly swap two people's order in the two-person graph when we transform the original data $C\times T\times 2\times N$ to $C\times T\times 2N$ tensor. The second solution is called symmetry processing. For each sample in the dataset, we create a reflected sequence by swapping two people's joint labels, i.e, we generate $C\times T\times 2\times 2N$ tensor instead of creating only one graph for each sample. When feeding into the network, we follow the idea of single-person GCN methods and split the 2 two-person skeletons as two samples. For example, suppose A and B represent two individuals with first $N$ joints and last $N$ joints, respectively. For a sample in the dataset "A is kicking B", we produce two sequences in the pre-processing phase, the original sequence "A is kicking B" and a reflected 'B is kicking A'. This operation is briefly explained in Figure~\ref{Fig:labeling}-(c).
    
\section{Experiments}
\subsection{Datasets}
    
    \paragraph{SBU Kinect Interactions}\cite{yun2012SBU} is a two-person skeleton interaction recognition dataset, containing eight human-human interaction actions (approaching, departing, pushing, kicking, punching, exchanging objects, hugging, and shaking hands). The SBU dataset has 282 short sequences, lasting 2-3 s each, involving seven different participants pairing up to 21 permutations. In each frame, 3D coordinates over 15 skeleton joints for each candidate are provided. We follow the 5-fold cross-validation evaluation defined by the authors and report the average accuracy.
    
    \paragraph{NTU-RGB+D and NTU-Interaction} NTU-RGB+D \cite{shahroudy2016ntu} is a general action recognition dataset containing both individual actions and two-person interactions. The full NTU-RGB+D dataset includes over 56,000 action sequences, covering 60 action classes (11 of them are interactions).  
    For interaction recognition, we only utilize the mutual interaction subset named NTU-Interaction, which is notwithstanding one of the largest skeleton interaction datasets up to now. NTU-Interaction contains 10,347 samples in 11 interactions. In each frame, 3D coordinates of 50 joints for two candidates are given. The authors proposed two widely-used benchmarks, and we follow them on NTU-Interactions. (1) \textbf{Cross-Subject (X-Sub)} divides the samples into training set (7,319 samples) and test set (3,028 samples) by subject id. (2) \textbf{Cross-View (X-View)} takes the camera ID as the criteria, in which samples taken by cameras 2 and 3 are used for training (6,889 samples), and samples of camera 1 are reserved for testing (3,458 samples). 
    
    
    \paragraph{NTU-RGB+D 120 and NTU120-Interaction}\cite{liu2019ntu120} is the extended version of NTU-RGB+D. NTU-RGB+D 120 expands NTU-RGB+D by adding another 57,367 samples of 60 novel action classes; 26 of the 120 classes are interactions. Similarly, we select the interaction subset denoted as NTU120-Interaction for interaction recognition.
    The authors also provide two recommended benchmarks: cross subject (\textbf{X-Sub120}) and cross setup (\textbf{X-Set120}). X-Sub120 takes 13,072 samples as the training set and 11,660 samples as the test set. In contrast, X-Set120 divides the samples by camera configurations: the distance and height of cameras. In X-Set120, the training set and test set have 11,864 and 12,868 samples, respectively.

\subsection{Implementation Details}
\label{sec:ImpDetails}
    
    \paragraph{Pre-processing}
    As discussed in section \ref{Sec:preprocessing}, the input features are formatted as $\mathbf{f}_{in}\in\mathbb{R}^{2C\times T\times 2N}$, where $C=3$ for 3D skeleton data. We resize each sample in NTU-Interaction to $T=64$ frames following \cite{chen2021CTR-GCN}. For SBU, we set $T=100$ and padding zeros after sequences less than $T$ frames. 
    
    \paragraph{Model Configurations}
    In input branches, the input-output channels for the three blocks are 6-64, 64-64, 64-32. In main branch, the channels for the six blocks are 128-128, 128-128, 128-128, 128-256, 256-256, 256-256. We utilize standard $3\times1$ temporal convolution for the first block in the input branches. 
    The batch size for the training set and test set is 16. We train our model for 65 epochs and use a warm-up strategy \cite{he2016deep} at the first 5 epochs. We choose cross-entropy as the loss function. The learning rate is initialized to 0.1, and we adopt an SGD optimizer with Nesterov momentum of 0.9 and weight decay of 0.0002. 
    All the experiments are conducted on one NVIDIA GeForce RTX 3090 GPU.

\subsection{Ablation Study for Interaction Recognition}

\begin{table}[tbp]
    \centering
    \caption{Ablation study on X-Sub120 of NTU120-Interaction dataset in FLOPs, parameter number, and averaged accuracy (\%) with standard error.} 
    \label{Table:Graph}
    \resizebox{\linewidth}{!}{
    \begin{tabular}{lcccc}
    \toprule
        Graph Scale         & Shape$(M,N)$ & FLOPs & \# Param. & Mean$\pm$Std. \\
    \midrule
        Baseline            & (2,25) & 1.87G & 1.47M & 91.28$\pm0.05$ \\
        Mutual              & (1,50) & 1.87G & 1.47M & 92.80$\pm0.10$ \\
        Random Swap         & (1,50) & 1.87G & 1.47M & 92.98$\pm0.07$ \\
        \textbf{Symmetry (Ours)}   & (2,50) & 3.74G & 1.47M & \textbf{93.47}$\pm0.05$ \\
    \midrule
    \specialrule{0em}{1.5pt}{1.5pt}
    \midrule
        Model Config       & Module            & FLOPs & \# Param. & Mean$\pm$Std. \\
    \midrule
        $3\times1$ Convolution & \multirow{2}*{TCN}  & 4.44G & 2.02M & 93.2$\pm0.08$ \\
        \textbf{Multi-scale (Ours)}     &           & 3.74G & 1.47M & \textbf{93.47}$\pm0.05$ \\        
                   
        PartAtt~\cite{song2020ResGCN}      & \multirow{3}*{Attention} & 3.74G & 1.87M & 92.83$\pm0.06$\\
        ST-JointAtt~\cite{song2022Eff-GCN} &        & 3.81G & 1.47M & 93.04$\pm0.20$ \\
        \textbf{ST-PartAtt (Ours)}      &           & 3.74G & 1.47M &  \textbf{93.47}$\pm0.05$ \\
    \midrule
    \specialrule{0em}{1.5pt}{1.5pt}
    \midrule
        Input data & Architecture & FLOPs & \# Param. & Mean$\pm$Std. \\
    \midrule
        Joint (J)           & \multirow{4}*{One-branch}  
                                            & \multirow{4}*{1.36G} 
                                                    & \multirow{4}*{1.31M} 
                                                            & 89.56$\pm0.10$ \\
        Bone (B)            &               &       &       & 90.12$\pm0.05$ \\
        Joint motion (JM)   &               &       &       & 87.35$\pm0.41$ \\     
        Bone motion (BM)    &               &       &       & 87.19$\pm0.01$ \\
        J + B               & Two-branch    & 1.55G & 1.37M & 91.24$\pm0.32$ \\
        J + B + JM          & Three-branch  & 1.74G & 1.43M & 92.19$\pm0.10$ \\
        \textbf{J + B + JM + BM (Ours)} 
                            & Four-branch   & 1.87G & 1.47M & \textbf{92.80}$\pm$\textbf{0.10} \\
    
    \bottomrule
    \end{tabular}
    }
\end{table}

    \paragraph{Graph Scale} 
    We first evaluate the effectiveness of our two-person graph representation for interaction recognition. The graph scale $(M,N)$ decides the number of graph nodes for each sample, where $M$ is the number of graphs created for each sample, and $N$ is the number of nodes in each graph. We compare the classification accuracies of our model under four graph scale configurations. (1) Baseline: The input two-person sequences are split into two separate skeletons, as most GCNs are adopted. (2) Mutual: Our two-person graph is adopted. (3) Random Swap: We randomly swap two people's relative order in the two-person graph. (4) Symmetry: This configuration enhances the input data by creating an additional reflected sequence for each sample. 
    The baseline is under a single-person graph, while configs 2-4 are based on the two-person graph with different processing techniques discussed in Section \ref{Sec:preprocessing}. We present the results in the top rows of Table \ref{Table:Graph}, where we have the following observations. First, the accuracy of our two-person graph significantly exceeds the baseline by over 1\% but has equal parameters and FLOPs. This result proves that our two-person graph has the advantage of the existing single-person graph representation in interaction recognition. Second, symmetry processing outperforms the vanilla two-person graph, which shows that symmetry processing can significantly improve the performance and address well the inherent asymmetry introduced by the two-person graph.
    
    \paragraph{Network Module}
    We compare the results of our model with those of two different temporal convolutional layers and three attention modules in the middle rows of Table \ref{Table:Graph}. Our model with Multi-scale TCN and with ST-Part Attention block has higher accuracy and lower or equal parameters than those with other TCN and attention blocks. This proves the effectiveness and robustness of our model.

    \paragraph{Multi-branch Input Data} 
    We feed four types of input data into four input branches and fuse them in the middle stage. To examine the necessity of each input branch, we present the experimental results with different input combinations under the Mutual graph scale $(1,50)$. As shown in the bottom rows in Table \ref{Table:Graph}, our model with all four input branches obtains the highest accuracy. As the number of branches increases, the accuracy improves but the computational costs (FLOPs) and parameters do not increase significantly. These results demonstrate that all four input data sets are informative for interaction recognition, and our model can effectively accumulate crucial information from these inputs.

\begin{table}[tbp]
    \centering
    \caption{Comparisons of different labeling strategies on NTU120-Interaction X-Sub120 benchmark in accuracy (\%).}
    \label{Table:Labelings}
    \begin{tabular}{c|c}
    \toprule
        Labeling strategy & Mean$\pm$Std. \\
    \midrule
        Fully-connected                 & 89.28$\pm0.23$ \\
        Only pairwise link              & 91.51$\pm0.04$ \\
        Pairwise labeling               & 92.04$\pm0.15$ \\
        Physical labeling               & 92.19$\pm0.18$ \\
        Interactive labeling            & 92.32$\pm0.09$ \\
        AAGC~\cite{shi2020AAGCN}        & 89.06$\pm0.17$ \\
        CTRGC~\cite{chen2021CTR-GCN}    & 89.99$\pm0.22$ \\
        \textbf{Geometric labeling (Ours)}       & \textbf{92.80}$\pm$\textbf{0.10} \\

    \bottomrule
    \end{tabular}

\end{table}

\begin{table}[tbp]
    \centering
    \caption{Comparisons between several GCNs with the two-person graph (with mutual suffix) and the single-person graph on NTU-Interaction and NTU120-Interaction in accuracy (\%).\\}
    \label{Table:GCNs-NTUmutual}
    \begin{tabular}{lcccc}
    \toprule
        \multirow{2}*{GCN} & \multicolumn{2}{c}{NTU-Interaction} & \multicolumn{2}{c}{NTU120-Interaction}\\ 
        ~ & X-Sub & X-View & X-Sub120 & X-Set120 \\
    \midrule
        ST-GCN\cite{yan2018ST-GCN} & 89.31 & 93.72 & 80.69 & 80.27 \\ 
        \textbf{ST-GCN mutual }
        & \textbf{91.05} & \textbf{94.58} & \textbf{83.39} & \textbf{82.46} \\  
        2S-AGCN\cite{shi20192sAGCN} & 93.36 & 96.67 & 87.83 & 89.21 \\
        \textbf{2S-AGCN mutual}
        & \textbf{93.96} & \textbf{97.22} & \textbf{87.93} & \textbf{89.69} \\ 
        Pa-ResGCN-B19\cite{song2020ResGCN} & 94.34 & 97.55 & 89.64 & 89.94 \\ 
        \textbf{Pa-ResGCN-B19 mutual}
        & \textbf{95.41} & \textbf{98.27} & \textbf{91.47} & \textbf{91.58} \\ 
        CTR-GCN\cite{chen2021CTR-GCN} & 96.33 & 98.75 & 92.03 & 92.82 \\ 
        \textbf{CTR-GCN mutual}
        & \textbf{96.76} & \textbf{98.87} & \textbf{92.26} & \textbf{92.91} \\ 
        EfficientGCN-B2\cite{song2022Eff-GCN} & 95.35 & 97.82 & 91.02 & 90.61 \\
        \textbf{EfficientGCN-B2 mutual}
        & \textbf{95.94} & \textbf{98.58} & \textbf{91.14} & \textbf{91.63} \\
    \bottomrule
    \end{tabular}
\end{table}

    \paragraph{Graph Labeling Strategy} 
    
    We compare our model's accuracy with those of different graph labeling strategies, as shown in Table \ref{Table:Labelings}. In geometric labeling, we average the input sequences of each batch and filter out correlations that are \textless0.3 for better graph sparsity. In Table~\ref{Table:Labelings}, we have the following observations: (1) A fully connected graph obtains the lowest accuracy, which shows that the skeletal structure is essential for classification. (2) Interactive labeling obtains higher performance than pairwise and physical labeling. These results prove the effectiveness of inter-body links and imply the importance of graph sparsity. (4) The introduced AAGC and CTRGC both dramatically reduce the accuracy, which indicates that the dynamic topology of a two-person graph does not always work as the single-person scenario. (5) Geometric labeling obtains the highest accuracy, showing that our model can effectively integrate intra-body and interactive information with a carefully designed graph structure.

\subsection{Effectiveness of the Two-person Graph Representation}

    To examine the effectiveness of the proposed two-person graph representation for convincing results, we evaluate several GCN models with our two-person graph. For each GCN, we only modify the graph scale to our two-person graph and keep other parameters and setups unchanged. As shown in Table \ref{Table:GCNs-NTUmutual}, all these GCNs with the two-person graph (with mutual suffix) obtain better performance than the original baseline. These results further prove the effectiveness and generalizability of our two-person graph in interaction recognition.

\begin{table}[tbp]
    \centering
    \caption{Comparisons between several GCNs with the two-person graph (with mutual suffix) and the single-person graph on the full NTU-RGB+D dataset in accuracy (\%).\\}
    \label{Table:GCNs-NTU}
    \begin{tabular}{lcccc}
    \toprule
        \multirow{2}*{GCN} & \multicolumn{2}{c}{NTU-RGB+D} & \multicolumn{2}{c}{NTU-RGB+D 120}\\ 
        ~ & X-Sub & X-View & X-Sub120 & X-Set120 \\
    \midrule
        ST-GCN\cite{yan2018ST-GCN} & 79.90 & 89.82 & 72.95 & 74.99 \\ 
        \textbf{ST-GCN mutual} & \textbf{81.65} & \textbf{90.96} & \textbf{75.41} & \textbf{76.15} \\ 
        Pa-ResGCN-B19 \cite{song2020ResGCN} & 90.82 & 95.78 & 86.55 & 87.94 \\
        \textbf{Pa-ResGCN-B19 mutual} & \textbf{91.05} & \textbf{95.98} & \textbf{86.96} & \textbf{88.43} \\
    \bottomrule
    \end{tabular}
\end{table}

\begin{figure}[tbp]
    \centering
    \includegraphics[width=0.48\textwidth]{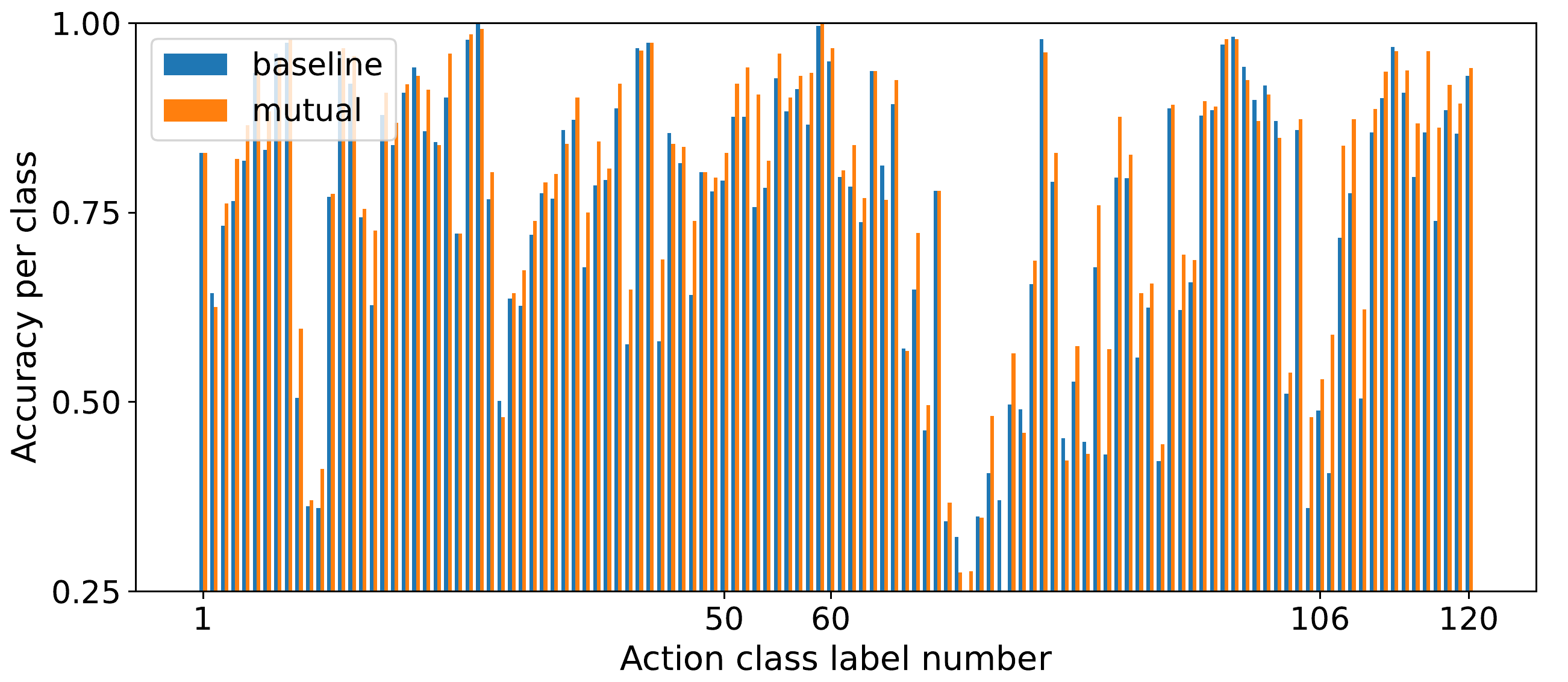}
    \caption{Accuracy comparison of each action class between ST-GCN baseline and ST-GCN with our two person graph on X-Sub120 benchmark of NTU-RGB+D 120 dataset. Actions of number 50-60, 106-120 are interactions.}
    \label{Fig:1Pvs2P}
\end{figure}
    
    Furthermore, the idea of two-person graph representation is compatible with single-person inputs. 
    The above experimental results show that our two-person graph retains the interactive information without additional computation costs. In addition, in order to investigate whether the two-person graph topology still works in general action recognition, we present the model performance of ST-GCN and Pa-ResGCN-B19 on the NTU-RGB+D full dataset in Table \ref{Table:GCNs-NTU}. When adopting a two-person graph with symmetry processing, ST-GCN gains significant improvement, and the accuracy of ResGCN still increases. To determine where 2P-GC outperforms 1P-GC, we compare the by-class accuracy on X-Sub120 of NTU-RGB+D 120 in Figure \ref{Fig:1Pvs2P}. We notice that ST-GCN with a two-person graph outperforms the baseline on almost all actions. The margins are more significant on interaction actions (no.50-60, 106-120). This is strong proof that the two-person graph representation is an ideal substitution for the original graph.

\subsection{Comparison with SOTA Methods}

\paragraph{SBU} 
    We compare the average accuracy of our best results on the 5-fold cross-validation with state-of-the-art results on SBU. We mainly compare our model with existing interaction recognition methods. As shown in Table \ref{Table:SBU}, the accuracy of our model is close to those obtained by SOTA methods. Note that SBU is a dataset containing only 282 samples. This experiment shows that our model can achieve promising results on a small dataset.

\begin{table*}[t]
    \centering
    \begin{threeparttable}
    \caption{Comparison with SOTA methods on NTU-Interaction and NTU120-Interaction dataset in accuracy (\%), FLOPs and parameter number.}
    \label{Table:NTU-Interaction}
    \begin{tabular}{lc|cccc|cc}
    \toprule
        \multirow{2}*{Method} & \multirow{2}*{Conf./Jour.} & \multicolumn{2}{c}{NTU-Interaction} & \multicolumn{2}{c}{NTU120-Interaction} & 
        \multirow{2}*{FLOPs} & \multirow{2}*{\# Param.} \\ 
        ~ & ~ & X-Sub & X-View & X-Sub120 & X-Set120 & ~ & ~ \\
    \midrule
        ST-LSTM$^*$ \cite{liu2017ST-LSTM}     
        & TPAMI17 & 83.00 & 87.30 & 63.00 & 66.60 & - & - \\
        GCA-LSTM$^*$ \cite{liu2017GCA-LSTM} 
        & CVPR17 & 85.90 & 89.00 & 70.60 & 73.70 & - & -\\
        2S-GCA-LSTM \cite{liu20172S-GCA-LSTM}                  
        & TIP17 & 87.20 & 89.90 & 73.00 & 73.30 & - & -\\
        ST-GCN-PAM \cite{yang2020pairwise}
        & ICIP20 & - & - & 83.28 & 88.36 & - & -\\
        LSTM-IRN \cite{perez2021LSTM-IRN}
        & TMM21 & 90.50 & 93.50 & 77.70 & 79.60 & - & -\\
        DR-GCN \cite{zhu2021DRGCN}
        & PR21 & 93.68 & 94.09 & 85.36 & 84.49 & 18.03G$^\Diamond$ & 3.31M$^\Diamond$ \\
        GeomNet \cite{nguyen2021geomnet}
        & ICCV21 & 93.62 & 96.32 & 86.49 & 87.58 & - & - \\
        K-GCN \cite{li2021K-GCN}
        & NEUROCOMP21 & 93.70 & 96.80 & - & - & - & - \\
        2S-DRAGCN \cite{zhu2021DRGCN}
        & PR21 & 94.68 & 97.19 & 90.56 & 90.43 & - & 7.14M$^\Diamond$\\
        AIGCN \cite{gao2022AIGCN}
        & ICME22 & 93.89 & 97.22 & 87.80 & 87.96 & - & 1.94M \\
        2s-AIGCN~\cite{gao2022AIGCN}
        & ICME22 & 95.34 & 98.00 & 90.71 & 90.65 & - & 3.88M \\
        \midrule
        ST-GCN$^\Diamond$ \cite{yan2018ST-GCN} 
        & AAAI18 & 89.31 & 93.72 & 80.69 & 80.27 & 16.32G & 3.10M \\
        AS-GCN$^*$ \cite{li2019AS-GCN}            
        & CVPR19 & 89.30 & 93.00 & 82.90 & 83.70 & 26.76G$^\Diamond$ & 9.50M$^\Diamond$ \\
        2S-AGCN$^\Diamond$ \cite{shi20192sAGCN}
        & CVPR19 & 93.36 & 96.67 & 87.83 & 89.21 & 37.32G & 6.94M \\
        Pa-ResGCN-B19$^\Diamond$ \cite{song2020ResGCN}
        & ACMMM20 & 94.34 & 97.55 & 89.64 & 89.94 & 17.78G & 3.64M \\
        CTR-GCN$^\Diamond$ \cite{chen2021CTR-GCN}
        & ICCV21 & 95.31 & 97.60 & 92.03 & 92.82 & 7.16G & 5.68M \\
        EfficientGCN-B2$^\Diamond$ \cite{song2022Eff-GCN}
        & TPAMI22 & 95.35 & 97.82 & 91.02 & 90.61 & 4.05G & \textbf{0.51M} \\
        \midrule
        \textbf{2P-GCN (Ours)}
        & - & \textbf{97.05} & \textbf{98.80} & \textbf{93.47} & \textbf{93.73}
        & \textbf{3.74G} & 1.47M \\
    \bottomrule
    \end{tabular}
    \begin{tablenotes}
    \item[*] Results are reported in \cite{perez2021LSTM-IRN}
    \item[$\Diamond$] We produce the results based on the code published by the authors.
    \end{tablenotes}
    \end{threeparttable}
\end{table*}

\begin{table}[t]
    \centering
    \caption{Comparisons of average accuracy (\%) with SOTA methods on SBU Interactions.}
    \label{Table:SBU}
    \begin{tabular}{lc}
        \toprule
           Method                                        & Accuracy \\
        \midrule
            ST-LSTM\cite{liu2017ST-LSTM}                 & 93.30 \\
            IRN$_{inter+intra}$\cite{perez2021LSTM-IRN}  & 96.10 \\
            K-GCN\cite{li2021K-GCN}                      & 97.20 \\
            LSTM-IRN$_{inter+intra}$\cite{perez2021LSTM-IRN} & 98.20  \\
            VA-fusion\cite{zhang2019VA-LSTM}             & 98.30  \\
            HCN\cite{li2018HCN}                          & 98.60  \\
            
            PJD+FCGC \cite{men2021two}                   & 96.80 \\
            GeomNet \cite{nguyen2021geomnet}             & 96.33 \\
            DR-GCN\cite{zhu2021DRGCN}                    & 99.06  \\
            AIGCN\cite{gao2022AIGCN}                     & \textbf{99.10} \\
            2P-GCN (Ours)                                & 98.90 \\ 
        \bottomrule
    \end{tabular}

\end{table}

\paragraph{NTU-Interaction and NTU120-Interaction} 
Table \ref{Table:NTU-Interaction} shows the comparison with the state-of-the-art methods. We compare 2P-GCN with two types of models. The first group consists of predominant methods in interaction recognition. Most of them are RNN-based methods~\cite{liu2017ST-LSTM,liu20172S-GCA-LSTM,perez2019IRN,perez2021LSTM-IRN} or GCN-based methods~\cite{yang2020pairwise,li2021K-GCN,zhu2021DRGCN,gao2022AIGCN}. These methods have been evaluated on NTU-Interactions, so the results of the first group are reported in the corresponding papers. 
The second group includes GCN-based methods in general skeleton-based action recognition~\cite{yan2018ST-GCN,li2019AS-GCN,shi20192sAGCN,song2020ResGCN,chen2021CTR-GCN,song2022Eff-GCN}. They were originally experimented on the full NTU-RGB+D dataset. We produce these results on NTU-Interaction and NTU120-Interaction using the code published by their authors with other settings unchanged.
For skeleton-based interaction recognition, our model outperforms current state-of-the-art method 2S-AIGCN~\cite{zhu2021DRGCN} by 1.71\%, 0.80\%, 2.76\% and 3.08\% on 4 benchmarks, respectively. Moreover, 2P-GCN is also superior to existing state-of-the-art method CTR-GCN~\cite{chen2021CTR-GCN} in skeleton-based general action recognition. Moreover, the proposed 2P-GCN brings a significant improvement in complexity with only 1.47M parameters and 3.74G FLOPs for a larger graph scale and four input branches.

\section{Conclusion}
    This paper proposes a two-person graph convolutional network for skeleton-based interaction recognition. First, we introduce our two-person graph as a substitution for the isolated single-person graph to represent human-human interactions. Further, we design two-person graph convolution block, which can effectively extract features of intra-body and inter-body interactions. When adopting our two-person graph topology, experiments showed significant accuracy improvement in both single-person and interaction actions for all tested GCNs. Experimental results on three datasets show that the proposed model obtains state-of-the-art performances. In the future, we will extend the proposed two-person graph to human-object interaction and group activity recognition and explore a generic representation of human actions.

\section{Acknowledgements}

    This work was supported by the Guangdong Basic and Applied Basic Research Foundation under Grant 2022A1515010800.


\bibliographystyle{IEEEtran}
\bibliography{conferences_abrv,references}{}

\end{document}